\documentclass{article}
\usepackage{spconf,amsmath,graphicx}
\usepackage{algorithmic}
\usepackage{array}
\usepackage{textcomp}
\usepackage{stfloats}
\usepackage{url}
\usepackage{verbatim}
\usepackage{algorithm}
\usepackage{graphicx}
\usepackage{subfigure}
\usepackage{graphicx}
\usepackage{multirow}
\usepackage{svg}
\usepackage{float}
\usepackage[figuresright]{rotating}
\usepackage{amsmath,amssymb,amsfonts}
\usepackage{mathtools}

\title{mini-PointNetPlus: A LOCAL FEATURE DESCRIPTOR IN DEEP LEARNING MODEL FOR 3D ENVIRONMENT PERCEPTION}
\name{Chuanyu Luo\textsuperscript{\rm 1, 2}, Nuo Cheng\textsuperscript{\rm 1}, Sikun Ma\textsuperscript{\rm 1}, Jun Xiang\textsuperscript{\rm 1}, Xiaohan Li\textsuperscript{\rm 1}, Shengguang Lei\textsuperscript{\rm 1}, Pu Li\textsuperscript{\rm 2}}
\address{
        \textsuperscript{\rm 1}LiangDao GmbH, 12099 Berlin, Germany\\
	\textsuperscript{\rm 2}Process Optimization Group, Technische Universität Ilmenau, 98693 Ilmenau, Germany\\
	\{chuanyu.luo, nuo.cheng, sikun.ma, jun.xiang, xiaohan.li, shengguang.lei\}@liangdao.de,  \{pu.li\}@tu-ilmenau.de}
\begin{document}
%
\maketitle
\begin{abstract}Common deep learning models for 3D environment perception often use pillarization/voxelization methods to convert point cloud data into pillars/voxels and then process it with a 2D/3D convolutional neural network (CNN). The pioneer work PointNet has been widely applied as a local feature descriptor, a fundamental component in deep learning  models for 3D perception, to extract features of a point cloud. This is achieved by using a symmetric max-pooling operator which provides unique pillar/voxel features. However, by ignoring most of the points, the max-pooling operator causes an information loss, which reduces the model performance. To address this issue, we propose a novel local feature descriptor, mini-PointNetPlus, as an alternative for plug-and-play to PointNet. Our basic idea is to separately project the data points to the individual features considered, each leading to a permutation invariant. Thus, the proposed descriptor transforms an unordered point cloud to a stable order. The vanilla PointNet is proved to be a special case of our mini-PointNetPlus. Due to fully utilizing the features by the proposed descriptor, we demonstrate in experiment a considerable performance improvement for 3D perception. 
\end{abstract}
\section{Introduction and related works}
\label{sec:intro}
Lidar is widely used for 3D environment perception in autonomous vehicles, robotics, and other applications. To detect and localize objects in 3D space poses a challenging task.  Unlike image-based computer vision, a point cloud from Lidar is unordered, sparse, and has varying density due to the distance information. To process an unordered point cloud, the pioneering work, PointNet~\cite{qi2017pointnet}, was proposed for directly operating on the raw point cloud for 3D indoor object classification and segmentation. Using a symmetric max-pooling layer, PointNet can handle unordered and irregular point clouds without losing the fine-grained information of objects.

Considering large-scale Lidar-based driving scenes, VoxelNet~\cite{zhou2018voxelnet} was proposed to divide an irregular point cloud into a set of dense 3D voxels, where a mini-PointNet is used as a voxel feature extraction module. And then a dense 3D-CNN is used as the backbone for object detection. After that, SECOND~\cite{yan2018second} was developed for real-time 3D object detection, in which a point cloud is divided into a set of sparse 3D voxels and thus a sparse 3D CNN on the non-empty voxels can be used. By projecting the 3D voxels to a 2D BEV feature map and 2D convolution, SECOND achieves high detection accuracy with a high speed. The mini-PointNet is also used in SECOND as a voxelwise feature extractor (VFE) module~\cite{zhou2018voxelnet}.  Besides, in a later version of SECOND, the mean value of the point cloud features is also represented as a voxel feature.

Inspired by PointNet and SECOND, PointPillars~\cite{lang2019pointpillars} was developed in which a point cloud is divided into 2D pillars as vertical columns of points. The features are also extracted by the mini-PointNet, on pillars instead of voxels. By only using 2D pillars with vertical spatial information, PointPillars achieves impressively high efficiency.

In the latest state-of-the-art approaches like ~\cite{fan2022SST}~\cite{yin2021center_point}~\cite{shi2022pillarnet}, the mini-PointNet has become a popular and standard component to extract unordered point cloud features on voxels/pillars. However, is the mini-PointNet really a perfect approach to feature extraction? The answer we think is no, since the non-parametric max-pooling used in it leads to much information loss by ignoring most points. In a similar case of 2D computer vision ~\cite{DB15a_convbetter}, it is reported that replacing the 2D max-pooling layer with a parametric strided 2D convolution can stabilize the model and improve the performance.

As far as we know, there have been very few studies which focus on extracting point cloud features for  large-scale driving scenes for autonomous driving. Like the vanilla PointNet, most works~\cite{qi2017pointnet++, zhao2021pointTF, li2018pointcnn} only focus on indoor classification and semantic segmentation from point clouds. A similar work is PASNet~\cite{cheng2022pasnet} which uses a learning-based method to sort an unordered point cloud. However, for $N$ points in a voxel/pillar, there is $N!$ permutation possibilities. Generally, all sorting-based methods map high dimensional data to a 1D sequence and thus can be sensitive to the number of points in voxels.

In this study, we consider high dimensional data in a point loud as a combination of the corresponding 1D dimensional data. Therefore, we project the data points to individual dimensions separately. As a result, instead of sorting high dimensional data, we sort the projected 1D dimension data, leading to a stable order. Finaly, a learning-based weight extract multiple features of the point cloud.

\section{Methodology}
\label{sec:format}

In this section, we present a novel grid feasture extractor (descriptor), mini-PointNetPlus, to aggregate the features of an unordered point cloud. Here, a grid means a pillar or a voxel. The proposed module can be used as a standard component in almost all 3D CNN for object detection in large-scale driving scenes. At first, the unordered point cloud problem will be stated and then the solution of vanilla PointNet analysed.  

Let $x \in \mathbb{R} ^ {C}$ denote a point with feature dimensions $C$. And $S = \{ x_1, x_2, ..., x_N \}$ denotes the point set including $N$ unordered points in a grid. We would like to find a solution $O  \in \mathbb{R} ^ {C}$, which actually describes the grid features, by an extractor function $O = f(x_1, x_2, ..., x_N)$. Since a point cloud is unordered, the output $O$ and function $f$ should be permutation invariant, which means
\begin{equation}
\label{eq:1}
\begin{aligned}
O \equiv f(x_1, x_2, ..., x_N) \equiv f(x_i, x_j, ..., x_k).
\end{aligned}
\end{equation}

Here $i, j, ..., k$ represent the permuted index set $\{1, 2, ..., N\}$. In PointNet, it is proposed that the extractor function $f$ is a combination of the multi-layer perceptron, and an invariant symmetric function by max-pooling, denoted as follows

\begin{equation}
\label{eq:2}
\begin{aligned}
&O = f(x_1, x_2, ..., x_N) = Max(h(x_1), h(x_2), ..., h(x_n)), \\
&Max : \mathbb{R} ^ {N \times C} \rightarrow \mathbb{R} ^ {C}
\end{aligned}
\end{equation}

Here $h$ is the multi-layer perceptron network. It is noted that the max-pooling function works on the same feature dimension across different points. Other symmetric functions include the addition, average pooling etc. Experiment results~\cite{qi2017pointnet} show that the max-pooling operator performs better than the other symmetric functions.

However, one obvious limitation of PointNet is that the max-pooling operator ignores the less important points. In our proposed approach, it can be proved that the symmetric function is in fact not the key to keep the permutation invariant property, and all the points can be considered with an adaptive weight.

\begin{figure}[H]
\begin{center}
\includegraphics[width=1.0\linewidth]{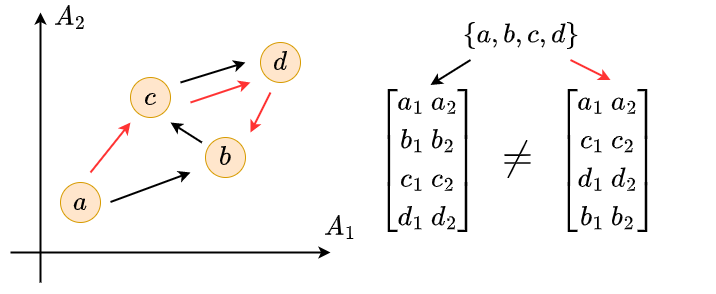}
\end{center}
\caption{For illustration purpose, only 2D points are shown. In practice, the points can be high dimensional data after multi-layer perceptron. The black and red arrows indicate two possible orders. There is actually no stable order to sort all the points by a function to map from set to matrix.}
\label{fig:sort_1}
\end{figure}

For $N$ points with dimension $C$, there exists $N!$ permutations, and there is no stable order to map the high dimensional data to 1D sequence. Fig.~\ref{fig:sort_1} shows two options of sorting paths. 

\begin{figure}[H]
\begin{center}
\includegraphics[width=1.0\linewidth]{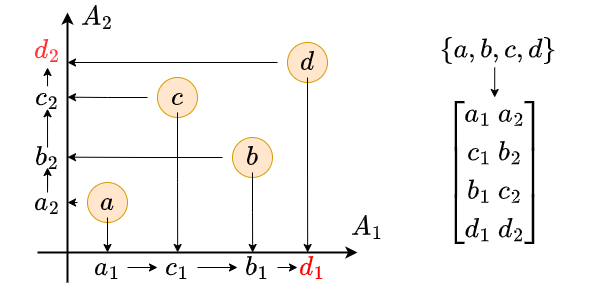}
\end{center}
\caption{2D points are projected to the $A_1$-axis and $A_2$-axis separately. There is a stable and permutation invariant order, i.e., to sort along the $A_1$-axis and $A_2$-axis, respectively. Note that PointNet considers only the red max values $d_1$ and $d_2$.}
\label{fig:sort_2}
\end{figure}

However, if projecting high dimensional points to each individual feature/channel dimension separately, there is a permutation invariant and a stable order in each dimension, as illustrated in Fig.~\ref{fig:sort_2}.

Therefore, for point set $S =\{ x_1, x_2, ..., x_N \}$, we define the set function in Eq.~\ref{eq:1} as $f(S) = Sort(Proj(S))$, where $f$ is a permutation invariant set function. The permutation invariant property of the set function can be expressed as Eq.~\ref{eq:proposed_func}, corresponding to Fig.~\ref{fig:sort_2}.

\begin{equation}
\label{eq:proposed_func}
\begin{aligned}
\begin{bmatrix}a_{1}  \ a_{2} \\ 
c_{1} \ b_{2} \\ 
b_{1} \ c_{2} \\ 
d_{1} \ d_{2} \end{bmatrix} \equiv f({a,b,c,d}) \equiv f({c,d,b,a}).
\end{aligned}
\end{equation}

Generally, the output $A \in \mathbb{R} ^ {N \times C}$ can be written as a matrix as follows

\begin{equation}
\label{eq:3}
\begin{aligned}
A &= f(h(x_1), h(x_2), ..., h(x_n))) \\
  & = \begin{bmatrix}
  a_{11} & \cdots & a_{1C}\\
 a_{21} & \cdots & a_{2C} \\
 \vdots & \ddots & \vdots \\
 a_{N1} & \cdots & a_{NC}
\end{bmatrix}.
\end{aligned}
\end{equation}

Here $f(S) = Sort(Proj(S))$. The $Proj$ function will project all points in each feature dimension and the $Sort$ function will sort the points ascendingly in each feature dimension. It means, the elements of any column index $j$ in $A$ always satisfy $a_{1j} < a_{2j} < \cdots < a_{Nj}$. The final output $O$ of the extractor function will be

\begin{equation}
\label{eq:4}
\begin{aligned}
O &= w^{T}A \\
  & = \begin{bmatrix}
  w_{1} & \cdots & w_{N}
  \end{bmatrix}
  \begin{bmatrix}
  a_{11} & \cdots & a_{1C}\\
 a_{21} & \cdots & a_{2C} \\
 \vdots & \ddots & \vdots \\
 a_{N1} & \cdots & a_{NC}.
\end{bmatrix}.
\end{aligned}
\end{equation}

Here $w^{T}$ is a vector of the learn-able weight parameters. Especially, in the vanillar PointNet, $w^{T} = \begin{bmatrix}
  0 & 0 & \cdots & 1
  \end{bmatrix}$.

\begin{table*}[ht!]
	\centering
        \resizebox{\textwidth}{!}{
	\begin{tabular}{|c|c|c|c|c|c|c|c|c|c|c|}
		\hline
		Method &\multicolumn{3}{|c|}{Car} & \multicolumn{3}{|c|}{Pedestrian} &\multicolumn{3}{|c|}{Cyclist} & speed\\
		\hline
		\multicolumn{1}{|c|}{ }& Easy & Moderate & hard & Easy & Moderate & hard & Easy & Moderate & hard& -\\
		\hline
		PointPillars~\cite{lang2019pointpillars} + PointNet~\cite{qi2017pointnet} & 85.17 & 76.29 & 74.22 & 56.66 & 52.38 & \textbf{47.78} & \textbf{80.69} & 62.74 & 58.97 & \textbf{22.3ms}  \\
            \hline
		PointPillars~\cite{lang2019pointpillars} + PASNet~\cite{cheng2022pasnet} & 86.40 & 76.62 & 74.15 & \textbf{59.20} &\textbf{53.13} & 47.77 & 78.18 & 60.73 & 57.32 & 25.5ms \\
		\hline
		PointPillars~\cite{lang2019pointpillars} + \textbf{miniPointNetPlus} & \textbf{87.62} & \textbf{77.63} & \textbf{75.49} & 57.72 & 52.06 & 47.19 & 79.13 & \textbf{62.88} & \textbf{60.14} & 23.5ms \\
  
		\hline
		\hline
		SECOND \cite{yan2018second} + Mean Point Features & 90.79 & 89.83 & 89.02 & 65.03 & \textbf{62.17} & 58.69 & 87.91 & 77.29 & 74.03 & 33.7ms \\
		\hline
		SECOND \cite{yan2018second} + PointNet~\cite{qi2017pointnet} & 90.75 & 89.79 & 88.91 & 63.48 &  60.99 & 58.41 & 86.71 & \textbf{78.90} & 74.56 & 34.8ms \\
		\hline
            SECOND \cite{yan2018second} + \textbf{miniPointNetPlus} & \textbf{90.82} & \textbf{89.92} & \textbf{89.08} & \textbf{65.11} &  61.62 & \textbf{59.00} & \textbf{90.18} & 77.02 & \textbf{74.58} & \textbf{33.0ms} \\
            \hline
	\end{tabular}}
        \caption{Results on KITTI Val 3D detection benchmark with AP. The speed is measured on a Tesla V100 with one GPU 32G.}
        \label{table:tab_1}
\end{table*}

\begin{table*}[ht!]

	\centering
        \resizebox{\textwidth}{!}{
	\begin{tabular}{|c|c|c|c|c|c|c|c|c|c|c|}
		\hline
		Method &\multicolumn{3}{|c|}{Car} & \multicolumn{3}{|c|}{Pedestrian} &\multicolumn{3}{|c|}{Cyclist} & speed\\
		\hline
		\multicolumn{1}{|c|}{ }& Easy & Moderate & hard & Easy & Moderate & hard & Easy & Moderate & hard& -\\
		\hline
		PointPillars~\cite{lang2019pointpillars} + PointNet~\cite{qi2017pointnet} & 90.72 & 89.39 & 88.28 & 45.53 & 42.34 & 40.30 & 84.53 & 71.70 & 68.25 & \textbf{22.3ms}  \\
            PointPillars~\cite{lang2019pointpillars} + PASNet~\cite{cheng2022pasnet} & 90.81 & 89.42 & 88.28 & \textbf{48.19} & \textbf{44.73} & \textbf{42.31} & 84.48 & 68.50 & 64.47 & 25.5ms \\
		\hline
		PointPillars~\cite{lang2019pointpillars} + \textbf{miniPointNetPlus} & \textbf{90.84} & \textbf{89.60} & \textbf{88.59} & 47.43 & 44.55 & 41.81 & \textbf{86.55} & \textbf{73.41} & \textbf{69.79} & 23.5ms \\
		\hline
		\hline
		SECOND \cite{yan2018second} + Mean Point Features & 88.31 & 78.37 & 77.22 & \textbf{57.84} & 53.72 & 48.11 & 80.88 & \textbf{66.85} & 62.58 & 33.7ms \\
		\hline
		SECOND \cite{yan2018second} + PointNet~\cite{qi2017pointnet} & 88.12 & \textbf{78.57} & \textbf{77.25} & 57.78 &  \textbf{53.90}& \textbf{49.17} & 80.15 & 65.82 & 61.38 & 34.8ms \\
		\hline
            SECOND \cite{yan2018second} + \textbf{miniPointNetPlus} & \textbf{88.38} & 78.40 & \textbf{77.25} & 57.54 &  53.64 & 48.82 & \textbf{82.78} & 66.83 & \textbf{62.91} & \textbf{33.0ms} \\
            \hline
	\end{tabular}}
        \caption{Results on KITTI Val 3D detection benchmark with AOS. The speed is measured on a Tesla V100 with one GPU 32G}
        \label{table:tab_2}
\end{table*}

\begin{table*}[ht!]

	\centering
        
	\begin{tabular}{|c|c|c|c|}
		\hline
		Method & mAP & NDS & speed\\
		\hline
		PointPillars~\cite{lang2019pointpillars} + PointNet~\cite{qi2017pointnet} & 44.07 & 57.71 & \textbf{22.6ms}  \\
            \hline
		PointPillars~\cite{lang2019pointpillars} + PASNet~\cite{cheng2022pasnet} & 41.73 & 56.41 & 27.5ms \\
		\hline
		PointPillars~\cite{lang2019pointpillars} + \textbf{miniPointNetPlus} & \textbf{44.51} & \textbf{58.18} & 22.7ms \\
            \hline
	\end{tabular}
        \caption{Results on nuScenes Val 3D detection benchmark. The speed is measured on a Tesla A100 with one GPU 40G}
        \label{table:tab_3}
\end{table*}

\section{Experiments}
\label{sec:pagestyle}

For a numerical verification, we take the pillar-based method PointPillars~\cite{lang2019pointpillars} and the voxel-based method SECOND~\cite{yan2018second} as our experiment pipeline. The mini-PointNet in the vanilla pipeline is replaced and compared with the adaptive sorting method PASNet~\cite{cheng2022pasnet} and the proposed  mini-PointNetPlus, respectively, for comparison. The KITTI~\cite{geiger2012kitti} and nuScenes~\cite{caesar2020nuscenes} dataset are employed as point cloud data in our experiment.

For the pillar-based method using PointPillars, the pillar size is set as $0.16m \times 0.16m \times 4m$ in KITTI and $0.2m \times 0.2m \times 8.0m$ in nuScenes, while the max points number in a pillar is taken as 32 in KITTI and 20 in nuScenes. For the voxel-based method using SECOND, the voxel size is set as $0.05m \times 0.05m \times 0.1m$  and the max points number in voxel is set as 5 in KITTI, respectively.

Except for the grid feature extraction module, all the other modules and training hyper-parameters are take with the same values.

In KITTI, the average precision (AP)  and the average orientation similarity (AOS)~\cite{geiger2012kitti} are used as evaluation metrics. In nuScenes, the evaluation metrics are AP and the nuScenes detection score (NDS)~\cite{caesar2020nuscenes}.

The results based on the KITTI validation set is listed in Tab.~\ref{table:tab_1} and Tab.~\ref{table:tab_2}, while the evaluation based on the nuScenes validation set is listed in Tab.~\ref{table:tab_3}.

It can be observed in the voxel-based method SECOND, all three voxel feature extraction modules have the similar performance. This is because that the voxel-based method takes only few points in each voxel (in the experiments the number is 5). The similar conclusion can also be found in PASNet~\cite{cheng2022pasnet}.

From the pillar-based method experiments of Tab.~\ref{table:tab_1} and Tab.~\ref{table:tab_2}, the adaptive sorting method PASNet~\cite{cheng2022pasnet} performs the best on the accuracy and orientation of pedestrian detection. The proposed method performs the best on the accuracy and orientation of car and cyclist detection.

In the nuScenes~\cite{caesar2020nuscenes} validation benchmark as shown in Tab.~\ref{table:tab_3}, the proposed method performs better than the vanilla PointNet and PASNet. Concerning the trade-off of accuracy and speed, our method has only a minor speed reduction compared to the PointNet baseline, but PASNet is significantly slower.

\section{Conclusion}

In this paper, we analyzed the mechanism and limitation of the widely applied method, PointNet, based on which we proposed an approach to separately projecting the individual features to 1D dimensions. It is proved that the PointNet is only one special case of our proposed mini-PointNetPlus. The symmetric function has been widely considered as the key solution to unordered points. But our method shows that if the points are projected separately in the projected dimension, we can find a stable and permutation invariant order. The symmetric function is only one handcrafted function to handle the projected points.

It is demonstrated by experiments that for pillar-based methods, in which pillar features can be fully represented by enough points, our proposed method provides a better performance with only a little cost of efficiency.

\bibliographystyle{IEEEbib}
\bibliography{refs}

\end{document}